\pdfoutput=1

\documentclass[11pt]{article}
\usepackage[final]{acl}
\usepackage{times}
\usepackage{latexsym}
\usepackage{makecell}

\usepackage[T1]{fontenc}
\usepackage[utf8]{inputenc}
\usepackage{microtype}
\usepackage{inconsolata}
\usepackage{graphicx}
\usepackage{algorithm}
\usepackage{algpseudocode}
\usepackage{amsmath}
\usepackage{mathtools} 
\usepackage{multirow}  

\title{From Teacher to Student: Tracking Memorization Through Model Distillation}

\author{
  Simardeep Singh \\
  Indian Institute of Technology, Roorkee \\
  Roorkee, Uttarakhand, India \\
  \texttt{simardeep\_s@mt.iitr.ac.in} \\
}

\begin{document}
\maketitle

\begin{abstract}
Large language models (LLMs) are known to memorize parts of their training data, raising important concerns around privacy and security. While previous research has focused on studying memorization in pre-trained models, much less is known about how knowledge distillation (KD) affects memorization. In this study, we explore how different KD methods influence the memorization of fine-tuned task data when a large teacher model is distilled into smaller student variants. This study demonstrates that distilling a larger teacher model, fine-tuned on a dataset, into a smaller variant not only lowers computational costs and model size but also significantly reduces the memorization risks compared to standard fine-tuning approaches.
\end{abstract}

\section{Introduction}
The rapid scaling of large language models (LLMs) has led to growing concerns about their ability to memorize and potentially reproduce sensitive training data. While earlier work has largely focused on describing memorization in LLMs through qualitative analysis \cite{carlini2021extractingtrainingdatalarge}, more recent research has introduced a quantifiable framework that evaluates memorization based on a model's ability to recall training examples verbatim when prompted \cite{carlini2023quantifyingmemorizationneurallanguage}, but crucially focused only on pre-trained models and their original training datasets. These foundational studies left open critical questions about memorization during fine-tuning -a common practice by which models are tuned to downstream tasks \cite{jiang2024improvingdomainadaptationextendedtext}. Fine-tuning is particularly risky because it tends to employ specialized, possibly sensitive data, e.g., medical records, proprietary data \cite{lukas2023analyzingleakagepersonallyidentifiable,kim2023propileprobingprivacyleakage,huang2022largepretrainedlanguagemodels}. In contrast to pre-training data that is generally broad and public, fine-tuning data is smaller and more specific, fine-tuning datasets are smaller and more targeted, making memorization both more likely and more dangerous. Addressing this gap, recent work by \cite{yang2024memorization} systematically investigates memorization and privacy risks in domain-specific LLMs. Their findings confirm that fine-tuned models, especially those trained on domain-specific corpora, are significantly prone to memorizing and potentially leaking sensitive content. Building on these findings, this study examines how knowledge distillation affects memorization in student models.

As large language models grow in capability and size, knowledge distillation (KD) \cite{hinton2015distillingknowledgeneuralnetwork} has emerged as a critical technique to reduce their computational demands, where we train a small student model with supervision from a large teacher model. This technique, which compresses knowledge from large "teacher" models into smaller "students", was originally developed for efficiency, its impact on memorization remains unexplored, particularly in the fine-tuning context. This study bridges this gap by systematically studying how different distillation methods affect memorization when transferring knowledge from a fine-tuned teacher to smaller students.

Our experiments demonstrate that distillation not only achieves its traditional benefits of reduced model size and computational costs, but also serves as an effective privacy-preserving technique by considerably decreasing memorization while preserving task performance. This dual advantage makes distillation particularly valuable for deploying LLMs in privacy-sensitive settings.

\section{Methodology}
\subsection{Defining Memorisation}
To study how memorization persists or changes through model distillation, we adopt a definition of memorization based on the framework introduced by \cite{carlini2023quantifyingmemorizationneurallanguage}, adapted for instruction-following tasks. Given an instruction-context-response tuple (p,c,s) from our fine-tuning dataset D, where p is the instruction (prefix),c is the context and s is the target response, we define:

\textbf{Memorization Criterion:} A model f is said to have memorized response s if, when prompted with instruction p and context c using greedy decoding, it generates s' such that:
\begin{itemize}
\item  s' exactly matches s (verbatim reproduction)
\item  The match persists for at least k tokens (k = length of s in our implementation)
\end{itemize}

The algorithm below formalizes the measurement of memorization over a dataset $\mathcal{D}$:

\begin{algorithm}
\caption{Memorization Measurement}\label{alg:memorization}
\begin{algorithmic}[1]
\Require Model $f$, dataset $\mathcal{D} = \{(p_i, c_i, s_i)\}_{i=1}^N$, threshold $k$
\Ensure Memorization fraction $M \in [0,1]$
\State $\text{memorized\_count} \gets 0$
\For{$(p, c, s) \in \mathcal{D}$}
    \State $\text{generated} \gets f(p, c, \text{max\_length}=\text{len}(s))$ \Comment{Greedy decoding with context}
    \If{$\text{exact\_match}(\text{generated}, s)$}
        \State $\text{memorized\_count} \gets \text{memorized\_count} + 1$
    \EndIf
\EndFor
\State \Return $\text{memorized\_count} / |\mathcal{D}|$
\end{algorithmic}
\end{algorithm}
\subsection{Distillation Methods}
\label{sec:Distillation-Methods}
To understand how memorization persists across student models, we apply four distinct knowledge distillation (KD) methods, each introducing different levels of supervision and approximation from the teacher to the student. The student models are trained on the same instruction-response data, but with guidance from the teacher model rather than the gold responses directly (except in SFT).
\subsubsection{Supervised Fine-Tuning (SFT)}
Supervised fine-tuning serves as a baseline method. The student model is directly trained on the ground-truth responses R, given the instruction I and context C, using the next-token loss objective. There is no teacher guidance in this process.
\begin{equation}
\mathcal{L}_{\text{SFT}} = - \sum_{t=1}^{T} \log P_\theta(R_t \mid R_{<t}, I,C)
\end{equation}
This approach represents direct learning from labeled data without model-to-model interaction.
\subsubsection{Word-Level Knowledge Distillation (WL-KD)}
Word-level knowledge distillation \cite{sanh2020distilbertdistilledversionbert,kim2016sequencelevelknowledgedistillation} involves training the student to mimic the teacher's token-level probability distribution over the vocabulary for each position in the output.
Let $\mathbf{s} = [s_1,\ldots,s_I]$ and $\mathbf{t} = [t_1,\ldots,t_J]$ be the student/teacher sentence, with $I$ and $J$ respectively being the their lengths.
\begin{multline}
\mathcal{L}_{\text{WORD-KD}} = - \sum_{j=1}^{J} \sum_{k=1}^{|V|} q(t_j = k \mid s, t_{<j}) \cdot \\
\log p(t_j = k \mid s, t_{<j})
\end{multline}

Here, 
q is the teacher's soft distribution, and p is the student's prediction. This formulation allows the student to receive richer supervision than hard targets, incorporating uncertainty and alternative possibilities. The student is further be trained to optimize the mixture of $\mathcal{L}_{\text{WORD-KD}}$ and $\mathcal{L}_{\text{WORD-NLL}}$.
\subsubsection{Sequence-Level Knowledge Distillation (Seq-KD)}
Sequence-level knowledge distillation \cite{kim2016sequencelevelknowledgedistillation} shifts from token-level supervision to entire sequence-level approximation. Instead of matching token probabilities, the student model attempts to match the full-sequence output generated by the teacher.

\begin{align}
\mathcal{L}_{\text{SEQ-KD}} 
&= - \log p(t = \hat{y} \mid s)
\end{align}

\noindent
Where $\hat{y} = \text{BeamSearch}(f_{\text{teacher}}, s)$ is the response sequence predicted by the teacher under beam search for a given instruction $I$.
\subsubsection{Reverse KLD Distillation (RKLD)}
The MiniLLM framework \cite{gu2024minillmknowledgedistillationlarge} proposes minimizing the reverse KL divergence from the student to the teacher, rather than the conventional forward KL used in KD.

\begin{equation}
\begin{split}
\theta &=\operatorname*{arg\,min}_\theta \text{KL}[q_\theta \| p] \\\end{split} 
\label{eq:rev_kl_instruction}
\end{equation}
We follow the MiniLLM \cite{gu2024minillmknowledgedistillationlarge} optimization procedure, where the model parameters $\theta$ are updated via:

\begin{equation}
\theta \leftarrow \theta - \eta \left[ {(\nabla\mathcal{L})_{\text{Single}} + (\nabla\mathcal{L})_{\text{Norm}}} + {\nabla\mathcal{L}_{\text{PT}}}_{} \right]
\end{equation}

until convergence, where:
\begin{itemize}
    \item $(\nabla\mathcal{L})_{\text{Single}}$ and $(\nabla\mathcal{L})_{\text{Norm}}$ compute the importance-weighted reverse KLD gradients
    \item $\nabla\mathcal{L}_{\text{PT}}$ maintains pretrained language model capabilities
    \item $\eta$ is the learning rate
\end{itemize}

\begin{table*}[t]
\centering
\begin{tabular}{l|l|l|l}
\hline
\textbf{Model} & \textbf{Params} & \textbf{Technique} & \makecell{\textbf{Fraction of} \\ \textbf{memorisation}} \\
\hline
\multirow{14}{*}{GPT2} & 1.5B & SFT & 0.654  \\
\cline{2-4}
& 760M & SFT & 0.523  \\
& 360M & SFT & 0.433  \\
& 120M & SFT & 0.330 \\
\cline{2-4}
& 760M & KD & 0.472  \\
& 360M & KD & 0.140  \\
& 120M & KD & 0.100 \\
\cline{2-4}
& 760M & SeqKD & 0.315  \\
& 360M & SeqKD & 0.134  \\
& 120M & SeqKD & 0.129  \\
\cline{2-4}
& 760M & RKLD & \textbf{0.090}  \\
& 360M & RKLD & \textbf{0.075}  \\
& 120M & RKLD & \textbf{0.060} \\
\hline
\end{tabular}
\caption{Fraction of memorisation across different GPT-2 model sizes and distillation techniques.}
\label{tab:memorisation-fraction}
\end{table*}

\subsection{Evaluation Criteria}
\label{sec:methods}
This study evaluates each student model trained via the above distillation methods using two metrics \textbf{Memorization Fraction} to measure verbatim copying using Algorithm~\ref{alg:memorization}  (detailed methodology provided in Section~\ref{sec:Experiments and Results})  and \textbf{ROUGE Scores} \cite{lin-2004-rouge}.
Although originally designed for summarization evaluation, we repurpose ROUGE metrics to analyze memorization behavior through different granularity levels by computing scores against the training targets and test targets. This reveals how closely generated responses replicate seen examples as well as generalize to unseen ones:

\begin{equation}
\text{ROUGE-N} = \frac{\sum_{S\in\mathcal{R}}\sum_{g_n\in S}C_{match}(g_n)}{\sum_{S\in\mathcal{R}}\sum_{g_n\in S}C(g_n)}
\end{equation}

where: $\mathcal{R}$ stands for Reference text sets, $g_n$ for $n$-gram and $C$ for count function.

For sequence-level analysis:

\begin{equation}
\text{ROUGE-L} = \frac{(1+\beta^2)R_\ell P_\ell}{R_\ell + \beta^2 P_\ell}
\end{equation}

with $R_\ell$ and $P_\ell$ being recall and precision of the longest common subsequence.

\section{Experiments and Results}
\label{sec:Experiments and Results}
\subsection{Experimental Setup}
In our experiments, we employ the GPT-2 family of models \cite{Radford2019LanguageMA} to evaluate memorization across various student-teacher configurations. For teacher model, we use GPT-2 1.5B and other three smaller variants GPT-2 760M, GPT-2 340M, and GPT-2 120M as student models.

We utilize the DollyEval (databricks-dolly-15k) dataset ($\mathcal{D}$), which contains instruction-response pairs curated to evaluate instruction-following capabilities. Due to computational limitations, the teacher model was fine-tuned on 10,000 examples from $\mathcal{D}$.From the remaining 5,000 examples, we randomly sampled 500 examples to evaluate ROUGE scores on test data .Subsequently, this fine-tuned teacher was used to distill knowledge into the student models using the techniques outlined in Section~\ref{sec:Distillation-Methods} — namely Supervised Fine-Tuning (SFT), Knowledge Distillation (KD), Sequence-level KD (SeqKD), and Reverse KL-based Distillation (RKLD).

As mentioned it section~\ref{sec:methods} ,we adopt two metrics to quantify memorisation in the models:

\begin{itemize}
    \item \textbf{Memorization Fraction}: 
    Using Algorithm~\ref{alg:memorization} with $k=50$ , we compute the fraction of verbatim reproductions from the training set across 3,000 randomly sampled examples. This metric directly quantifies the extent of data memorization.
    
    \item \textbf{ROUGE Scores}: 
    We calculate average ROUGE scores over the 500 examples in both the train and test dataset, between the generated and original responses. High scores on the training set combined with high memorization fractions suggest verbatim copying of training examples.
\end{itemize}

\subsection{Results}
Table~\ref{tab:memorisation-fraction} reports the fraction of memorization for distilled GPT-2 models across sizes and distillation techniques. The results reveal some interesting patterns. Larger models consistently exhibit higher memorization, confirming the correlation between capacity and verbatim recall. SFT, which directly fine-tunes the model on the dataset without teacher guidance, resulted in the highest memorization fraction and the highest ROUGE scores on the training set suggesting that SFT encourages direct pattern memorization, leading to inflated n-gram overlap on the training data (Tables~\ref{tab:tab2}--\ref{tab:tab3}--\ref{tab:tab4})

In contrast, distillation methods showed lower memorization fractions and more balanced ROUGE scores between train and test sets. For instance, MiniLLM (RKLD) achieved the lowest memorization overall (0.065 for 120M,0.075 for 340M, 0.090 for 760M) while maintaining reasonable test set ROUGE scores. This is in accordance with its goal of minimizing reverse KL divergence, which penalizes overconfidence on training examples by discouraging memorization by design. Notably, KD and SeqKD also demonstrated lower memorization and train ROUGE scores than SFT while achieving comparable test-set ROUGE performance.

These results supports our argument that distillation is a more principled method for training deployable models on sensitive datasets as it not only offers lower computational costs and faster inference but reduces memorization risks.

\section{Limitations and Future Work}
This study is limited by the use of only a single dataset (DollyEval) due to computational constraints,which may not fully capture diverse memorization behaviors across tasks. The memorization analysis was conducted with a fixed 50-token window (for the memorisation fraction part), while examining varying sequence lengths could yield more comprehensive insights. Future work should validate these results across different model architectures beyond GPT-2 and incorporate additional metrics like perplexity and BLEU for a more complete evaluation of model behavior and memorization patterns. 

\section{Conclusion}
This study establishes knowledge distillation as a powerful technique for addressing the privacy challenges of deploying large language models. Our findings reveal that distillation not only fulfills its original objective of model compression and computational efficiency, but also plays a critical role in mitigating the privacy risks posed by memorization. While larger models and direct fine-tuning were associated with higher memorization and inflated training set ROUGE scores,the student models produced via distillation consistently demonstrated lower memorization fractions. 
\vspace{2\baselineskip}  

These insights suggest that distillation offers a dual benefit by enabling deployability in a resource constrained settings while simultaneously enhancing privacy by reducing a model’s tendency to memorize confidential or sensitive data. Future work should explore how to optimize the privacy-utility tradeoff further and extend this to several other architectures and domains.

\section{Ethical Considerations}
This work addresses a critical issue in the ethical deployment of language models—the risk of memorizing and inadvertently leaking sensitive information from training data. By exploring the privacy-preserving potential of knowledge distillation, we aim to contribute toward safer, more responsible AI development practices.
While our results suggest that KD reduces verbatim memorization, it does not guarantee complete privacy protection. Distilled models may still exhibit forms of implicit memorization or generalization that could be exploited by more sophisticated extraction techniques.
Second, our evaluation of memorization does not account for biases, toxicity, or fairness issues that may also propagate through the distillation process. These factors, while not the focus of our current study, are equally important in the context of safe and ethical model deployment.
In summary, while our findings point to promising directions for mitigating privacy risks through distillation, ethical deployment requires a multi-faceted approach involving evaluation of bias, robustness, and formal privacy metrics in addition to memorization.
\begin{figure}[ht]
\centering

\begin{minipage}{\linewidth}
\centering
\begin{tabular}{l|l|l|c c}
\hline
\multirow{2}{*} {Model} & \multirow{2}{*}{Params} & \multirow{2}{*}{Technique} & \multicolumn{2}{c}{R-1} \\
 & & & {Train} & {Test} \\
\hline
\multirow{14}{*}{GPT2} & 1.5B & SFT (Teacher) & 0.88 & 0.33 \\ 
\cline{2-5}
 & 760M & SFT & 0.78 & 0.31 \\
 & 340M & SFT & 0.72 & 0.30 \\
 & 120M & SFT & 0.67 & 0.25 \\
\cline{2-5}
 & 760M & KD & 0.73 & 0.34 \\
 & 340M & KD & 0.58 & 0.29 \\
 & 120M & KD & 0.65 & 0.28 \\
\cline{2-5}
 & 760M & SeqKD & 0.69 & 0.32 \\
 & 340M & SeqKD & 0.58 & 0.30 \\
 & 120M & SeqKD & 0.75 & 0.27 \\
\cline{2-5}
 & 760M & RKLD & 0.45 & 0.36 \\
 & 340M & RKLD & 0.57 & 0.34 \\
 & 120M & RKLD & 0.46 & 0.30 \\
\hline
\end{tabular}
\captionof{table}{ROUGE-1 scores across model sizes and techniques}
\label{tab:tab2}
\end{minipage}

\vspace{1mm} 

\begin{minipage}{\linewidth}
\centering
\begin{tabular}{l|l|l|c c}
\hline
\multirow{2}{*} {Model} & \multirow{2}{*}{Params} & \multirow{2}{*}{Technique} & \multicolumn{2}{c}{R-2} \\
 & & & {Train} & {Test} \\
\hline
\multirow{14}{*}{GPT2} & 1.5B & SFT (Teacher) & 0.85 & 0.14 \\ 
\cline{2-5}
 & 760M & SFT & 0.70 & 0.12 \\
 & 340M & SFT & 0.80 & 0.12 \\
 & 120M & SFT & 0.73 & 0.11 \\
\cline{2-5}
 & 760M & KD & 0.71 & 0.16 \\
 & 340M & KD & 0.44 & 0.12 \\
 & 120M & KD & 0.53 & 0.11 \\
\cline{2-5}
 & 760M & SeqKD & 0.60 & 0.14 \\
 & 340M & SeqKD & 0.43 & 0.11 \\
 & 120M & SeqKD & 0.65 & 0.10 \\
\cline{2-5}
 & 760M & RKLD & 0.39 & 0.16 \\
 & 340M & RKLD & 0.42 & 0.14 \\
 & 120M & RKLD & 0.29 & 0.13 \\
\hline
\end{tabular}
\captionof{table}{ROUGE-2 scores across model sizes and techniques}
\label{tab:tab3}
\end{minipage}

\end{figure}

\begin{table}[H]
\centering
\begin{tabular}{l|l|l|c c}
\hline
\multirow{2}{*} {Model} & \multirow{2}{*}{Params} & \multirow{2}{*}{Technique} & \multicolumn{2}{c}{R-L} \\
 & & & {Train} & {Test} \\
\hline
\multirow{14}{*}{GPT2} & 1.5B & SFT (Teacher) & 0.78 & 0.27 \\ 
\cline{2-5}
 & 760M & SFT & 0.76 & 0.25 \\
 & 340M & SFT & 0.76 & 0.25 \\
 & 120M & SFT & 0.66 & 0.24 \\
\cline{2-5}
 & 760M & KD & 0.72 & 0.29 \\
 & 340M & KD & 0.54 & 0.25 \\
 & 120M & KD & 0.62 & 0.24 \\
\cline{2-5}
 & 760M & SeqKD & 0.72 & 0.26 \\
 & 340M & SeqKD & 0.54 & 0.24 \\
 & 120M & SeqKD & 0.74 & 0.22 \\
\cline{2-5}
 & 760M & RKLD & 0.40 & 0.30 \\
 & 340M & RKLD & 0.53 & 0.28 \\
 & 120M & RKLD & 0.42 & 0.21 \\
\hline
\end{tabular}
\caption{ROUGE-L scores across model sizes and techniques}
\label{tab:tab4}
\end{table}
\newpage
\bibliography{latex/custom}
\end{document}